\documentclass[conference]{IEEEtran}
\IEEEoverridecommandlockouts
\usepackage{cite}
\usepackage{amsmath,amssymb,amsfonts}
\usepackage{enumerate}
\usepackage[ruled,norelsize]{algorithm2e}
\usepackage{multirow}
\usepackage{float}

\makeatletter
\newcommand{\removelatexerror}{\let\@latex@error\@gobble}
\makeatother
\usepackage{stfloats}
\usepackage{graphicx}
\usepackage{textcomp}
\usepackage{xcolor}
\def\BibTeX{{\rm B\kern-.05em{\sc i\kern-.025em b}\kern-.08em
    T\kern-.1667em\lower.7ex\hbox{E}\kern-.125emX}}
\graphicspath { {./images/} }
\usepackage{amssymb}
\usepackage{pifont}
\newcommand{\cmark}{\ding{51}}
\newcommand{\xmark}{\ding{55}}
\begin{document}

\title{Learning-based AC-OPF Solvers on Realistic Network and Realistic Loads}

\author{\IEEEauthorblockN{Tsun Ho Aaron Cheung\IEEEauthorrefmark{1},
Min Zhou\IEEEauthorrefmark{2}\IEEEauthorrefmark{3} and Minghua Chen\IEEEauthorrefmark{2}\IEEEauthorrefmark{4}}
\IEEEauthorblockA{\IEEEauthorrefmark{2}School of Data Science, City University of Hong Kong, HKSAR, China\\
Email: \IEEEauthorrefmark{1}aaronthcheung@gmail.com,
\IEEEauthorrefmark{3}mzhou.cityu@gmail.com,
\IEEEauthorrefmark{4}minghua.chen@cityu.edu.hk}}

\maketitle

\begin{abstract}
Deep learning approaches for the Alternating Current-Optimal Power Flow (AC-OPF) problem are under active research in recent years. A common shortcoming in this area of research is the lack of a dataset that includes both a realistic power network topology and the corresponding realistic loads. To address this issue, we construct an AC-OPF formulation-ready dataset called TAS-97 that contains realistic network information and realistic bus loads from Tasmania’s electricity network. We found that the realistic loads in Tasmania are correlated between buses and they show signs of an underlying multivariate normal distribution. Feasibility-optimized end-to-end deep neural network models are trained and tested on the constructed dataset. Trained on samples with bus loads generated from a fitted multivariate normal distribution, our learning-based AC-OPF solver achieves 0.13\% cost optimality gap, 99.73\% feasibility rate, and 38.62 times of speedup on realistic testing samples when compared to PYPOWER.
\end{abstract}


\section{Introduction}
The Alternating Current-Optimal Power Flow (AC-OPF) problem aims to minimize the generator dispatch costs while meeting the load demands in an AC power network and satisfying physical and network constraints. It is a problem faced daily by power companies that, if solved reliably in near real-time, could result in tremendous savings in operation costs. The non-convexity of this optimization problem prevents any quick solutions using conventional optimization methods. Despite the difficulty, researchers and industry practitioners have been actively searching for better solutions to the problem in the last decades.
Recently, deep learning methods have shown promising results to solve AC-OPF problems with their capability to quickly map network loads to dispatch solutions. Several literatures have shown neural networks can produce similar solutions to conventional solvers in a fraction of the conventional computation time. Most literatures perform numerical experiments on an existing power network. However, the network loads used for testing in their experiments are sampled from estimated probability distributions which could be a poor representation of real network loads. Whether learning-based solvers are capable of real network topology and its corresponding real network loads remains a question, of which a solid answer is needed for the industry to better evaluate their effectiveness.

We summarize our main contributions in the following. First, we integrate multiple data sources into one coherent realistic dataset that contains all the required information for AC-OPF formulation. The constructed dataset is an AC-OPF formulation-ready dataset that represents a realistic network topology and includes months of the corresponding realistic bus load data with a short sample time interval. Second, we perform numerical experiments to show that given training samples that have a close resemblance to the realistic testing samples, learning-based AC-OPF solvers can achieve a small optimality gap, high feasibility rate, and two orders of magnitude faster than a conventional AC-OPF solver.

\section{The AC-OPF Formulation}

Define the following symbols:
\begin{tabular}{lll}
&\emph{Symbol}&\emph{Definition}\\
&\(\mathcal{N}\)&Set of buses\\
&\(\mathcal{G}\)&Set of P-V buses\\
&\(\mathcal{D}\)&Set of P-Q buses\\
&\(\mathcal{E}\)&Set of branches\\
&\(P_{Gi}\)&Active power generation on bus \(i\), \(i\in\mathcal{N}\)\\
&\(P_{Di}\)&Active power load on bus \(i\), \(i\in\mathcal{N}\)\\
&\(Q_{Gi}\)&Reactive power generation on bus \(i\), \(i\in\mathcal{N}\)\\
&\(Q_{Di}\)&Reactive power load on bus \(i\), \(i\in\mathcal{N}\)\\
&\(V_{i}\)&Complex voltage on bus \(i\), \(i\in\mathcal{N}\)\\
&\(X^{\min}\)&Minimum operation limit of quantity \(X\)\\
&\(X^{\max}\)&Maximum operation limit of quantity \(X\)\\
&$S_{ij}$&Apparent power on branch $(i,j)\in\mathcal{E}$ \\
&$y_{ij}$&Complex admittance on branch $(i,j)\in\mathcal{E}$\\
&Re\{\(z\)\}&Real part of complex number \(z\)\\
&Im\{\(z\)\}&Imaginary part of complex number \(z\)\\
&\(z^*\)&Conjugate of complex number \(z\)\\
\end{tabular}
\\

The AC-OPF problem can be formulated as follows \cite{Pan:2021}:
\begin{align}
\min  \quad  &\sum\nolimits_{i\in\mathcal{N}}{C_{i}(P_{Gi})},\\
\text{s.t.} \quad    &\sum\nolimits_{(i,j)\in\mathcal{E}}{\text{Re}\{V_{i}(V_{i}^{*}-V_{j}^{*})\}} = P_{Gi}-P_{Di}, \\
                   &\sum\nolimits_{(i,j)\in\mathcal{E}}{\text{Im}\{V_{i}(V_{i}^{*}-V_{j}^{*})\}} = Q_{Gi}-Q_{Di},  \\  
                   &P_{Gi}^{\min} \leq P_{Gi} \leq P_{Gi}^{\max}, i \in \mathcal{N}, \\
		&Q_{Gi}^{\min} \leq Q_{Gi} \leq Q_{Gi}^{\max}, i \in \mathcal{N}, \\
		&\lvert V_{i}\rvert^{\min} \leq \lvert V_{i} \rvert \leq \lvert V_{i}\rvert^{\max}, i \in \mathcal{N}, \\
		&\lvert V_{i}(V_{i}^{*}-V_{j}^{*})y_{ij}^{*}\rvert \leq S_{ij}^{\max}, (i,j)\in \mathcal{E}, \\
\text{var.} \quad    &P_{Gi},Q_{Gi},V_i,i\in \mathcal{N} \nonumber.		
\end{align}

Equation (1) is the total generation cost that we want to minimize. Equations (2) and (3) are the power balance equations. Equations (4) to (7) are the operation constraints of the buses and transmission lines. This optimization problem is nonconvex because Equations (2), (3) and (7) are nonconvex constraints.

\section{Literature Review}

\subsection{Research of learning-based AC-OPF solvers}
\begin{figure}[h!]
\centerline{\includegraphics[width=8cm]{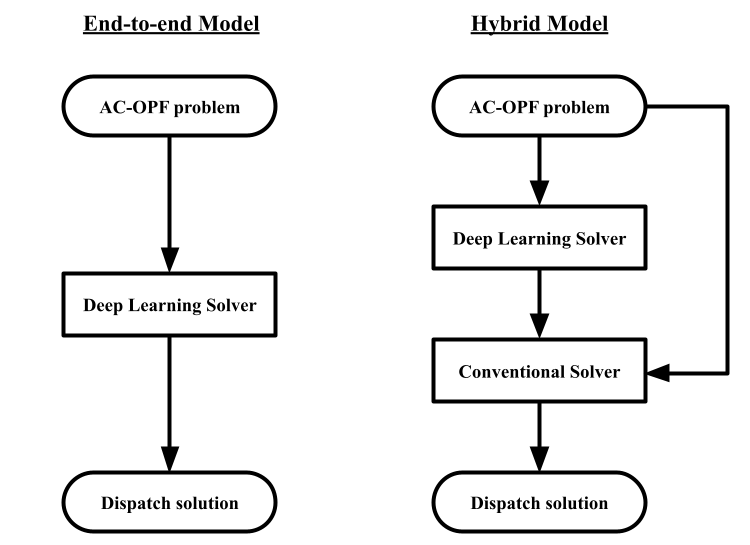}}
\caption{Flowchart of end-to-end model and hybrid model}
\label{figE2EandHybrid}
\end{figure}

\cite{Falconer:2021} categorizes learning-based AC-OPF solvers into either end-to-end models where mapping to OPF solutions are directly learned, or hybrid models where the neural network outputs are used as inputs to a conventional optimization solver. Fig.~\ref{figE2EandHybrid} shows the flowchart of the two models. According to its survey, end-to-end models have relatively short prediction time but high constraint violation rates, while hybrid models have relatively long prediction time but low constraint violation rates.

Works from \cite{Falconer:2021, Pan:2021,Mak:2021,Nellikkath:2021,Chatzos:2021} adopt the end-to-end approach. \cite{Pan:2021} tackles the feasibility drawback of end-to-end models by introducing constraint violation loss to the objective function in training. \cite{Falconer:2021} attempts to leverage local information from the power grid topology by adopting a learning-based model with a graph neural network. \cite{Mak:2021} improves the scalability of learning-based AC-OPF solvers by reducing the problem size through load embedding. \cite{Nellikkath:2021} proposes a physics-informed neural network that is capable of training on collocation points without traditional OPF ground truths. \cite{Chatzos:2021} improves the scalability of learning-based AC-OPF solvers by dividing a power network into islands and using a smaller neural network on each island.

Works from \cite{Venzke:2021,Chen:2020,Deka:2019} adopt the hybrid approach. \cite{Venzke:2021} improves the practicality of learning-based AC-OPF solvers in a safety-critical setting by developing a neural network that can classify the safeness of dispatch setpoints. \cite{Chen:2020} uses a 1-D convolutional neural network to output the voltage and phase angle as "hot-start" conditions for a conventional AC-OPF solver. \cite{Deka:2019} classifies the active set of constraints at optimality with which simple manipulations give the DC-OPF solutions.

\subsection{Research Limitation}
To facilitate discussion in this paper, we define the following terms. Realistic networks are networks that are a representation of an existing power network. Realistic loads are historical loads in a realistic network. Realistic samples are samples that contain realistic loads. Synthetic loads are bus loads that are generated by sampling from estimated distributions. Synthetic samples are samples that contain synthetic loads.

\cite{Falconer:2021, Pan:2021, Mak:2021, Nellikkath:2021,Chatzos:2021,Venzke:2021,Chen:2020,Deka:2019} all perform numerical experiments on realistic networks. However, the testing dataset used in their experiment are synthetic samples. \cite{Falconer:2021, Pan:2021, Mak:2021, Chatzos:2021} use uniform distribution with mean set as one reference load data to generate synthetic loads. \cite{Nellikkath:2021, Venzke:2021} use the latin hypercube sampling method \cite {McKay:1979} for synthetic sampling while \cite{ Chen:2020,Deka:2019} use the normal distribution.

Regardless of the distribution family used in generating the synthetic samples, \cite{Falconer:2021, Pan:2021, Mak:2021, Nellikkath:2021, Venzke:2021,Chen:2020,Deka:2019} share the same two limitations. First, they estimate the distribution parameters from a limited sample size, therefore the distribution parameter estimates may be inaccurate. Second, they have assumed the distributions of bus loads are independent of each other, which is not a necessarily true property of realistic loads. \cite{Chatzos:2021} removes the assumption of independent bus loads by including shared geographical and weather factors among nearby buses. Nonetheless, it still estimates the mean parameter of the uniform distribution for bus load from only one reference bus load data.

Because of the above limitations, the testing perofmrance from \cite{Falconer:2021, Pan:2021, Mak:2021, Nellikkath:2021,Chatzos:2021,Venzke:2021,Chen:2020,Deka:2019} on synthetic testing dataset may not reflect the model performance on realistsic samples. Our work directly uses a realistic testing dataset to avoid these limitations. Table~\ref{tabLiteratureReview} shows a comparison between different works.

\begin{table}[htbp]
\caption{Summary of literature review}
\begin{center}
\begin{tabular}{|c|c|c|c|c|}
\hline
\textbf{Literature}&\shortstack[l]{\textbf{Realistic}\\ \textbf{network}}& \shortstack[l]{\textbf{Testing}\\ \textbf{samples}} & \shortstack[l]{\textbf{Distribution}\\ \textbf{estimated} \\ \textbf{from $>20$ samples}} & \shortstack[l]{\textbf{Bus load} \\ \textbf{correlation}}\\
\hline
\shortstack[l]{\cite{Falconer:2021, Pan:2021, Mak:2021, Nellikkath:2021} \\ \cite{Venzke:2021,Chen:2020,Deka:2019}} & \cmark & synthetic & \xmark&\xmark\\
\hline
\cite{Chatzos:2021}& \cmark & synthetic & \xmark &\cmark\\
\hline
This work& \cmark & realistic & - &- \\
\hline
\end{tabular}
\label{tabLiteratureReview}
\end{center}
\end{table}

\section{Data sources for dataset construction}

\subsection{Realistic loads}
Historical substation loads from Australia’s National Electricity Market (NEM) 960-bus power network is publicly available\cite{Ausgrid:load, Endeavour:load, Essential:load, Evoenergy:load,CitiPower:load, Powercor:load, Jemena:load, United:load,Energex:load, Ergon:load,SAPN:load, TasNetworks:load}. The NEM network interconnects five regional electricity market jurisdictions (NEM regions) in Australia – Queensland, New South Wales (including the Australian Capital Territory), Victoria, South Australia, and Tasmania. Table~\ref{tabLoadDataSummary} shows a summary of the available load data in the NEM region.

\begin{table}[htbp]
\caption{NEM Load Data Summary}
\begin{center}
\begin{tabular}{|c|c|c|c|c|c|}
\hline
\textbf{NEM}&\textbf{Company}&\multicolumn{2}{|c|}{\textbf{Load Type}}&\multicolumn{2}{|c|}{\textbf{Period}} \\
\cline{3-6} 
\textbf{Region} &\textbf{Name} & \textbf{\textit{Active}}& \textbf{\textit{Reactive}}& \textbf{\textit{From}}&\textbf{\textit{To}$^{\mathrm{*}}$} \\
\hline
NSW& Ausgrid&\cmark &\xmark & 2005&2021\\
\hline
NSW& Endeavour&\cmark &\xmark & 2010&2020\\
\hline
NSW& Essential Energy&\cmark &\xmark & 2005&2018\\
\hline
NSW& Evoenergy&\cmark &\xmark & 2004&2021\\
\hline
VIC& CitiPower&\cmark &\cmark & 2010&2020\\
\hline
VIC& Powercor&\cmark &\cmark & 2010&2020\\
\hline
VIC& Jemena&\cmark &\cmark & 2008&2019\\
\hline
VIC& United Energy&\cmark &\xmark & 2011&2020\\
\hline
QLD& Energex&\cmark &\cmark & 2012&2020\\
\hline
QLD& Ergon&\cmark &\cmark & 2012&2021\\
\hline
SA& SAPN&\cmark &\cmark & 2012&2021\\
\hline
TAS& Tasnet&\cmark &\cmark & 2004&2018\\
\hline
\multicolumn{6}{l}{$^{\mathrm{*}}$As of 1\textsuperscript{st} March 2022}
\end{tabular}
\label{tabLoadDataSummary}
\end{center}
\end{table}

Only 7 out of 12 distribution companies in the NEM region have provided both active and reactive substation loads. There is insufficient real load data for AC-OPF study for the NEM network as a whole. To ensure our data’s truthfulness to reality, we require a part of the NEM network where all its distribution companies have provided both active and reactive load data, and whose connection to the rest of NEM network can be justifiably neglected. QLD, SA, and TAS networks all satisfy the first requirement. The TAS network also satisfies the second requirement in the period from 21\textsuperscript{st} December 2015 to 13\textsuperscript{rd} June 2016 because its high-voltage direct current interconnector to the Australian mainland was disconnected in that period due to a hardware problem. Therefore, realistic load data in the TAS network \cite{TasNetworks:load} in the disconnected period is used to construct our dataset.

\subsection{Realistic network topology}
Dataset \cite{Xenophon:2018} includes topological information of the TAS network, e.g. minimum dispatch and registered capacity of generators, resistance and reactance and susceptance of transmission lines, location and voltage of buses, etc.. This dataset would be used to formulate the constraints in Equations (4) and (6). $200MW$ is added to P-V bus 676's active generation capacity because of the temporary capacity boosting implemention for dealing with the crisis of interconnector outage \cite{Diesel:2016}. The generation cost function for each generator required in Equation (1), the reactive generation limits required in Equation (5), as well as the branch flow limits required in Equation (7) are not publicly available. Their estimation would be explained in the next section.

\section{Construction of TAS-97 dataset}
Each power station and substation is simplified as one node \cite{Xenophon:2018}, which translates to the 97 buses in our dataset. The constructed dataset is named TAS-97 to indicate the real-world network it represents and the number of buses it contains. 

\subsection{Assigning loads to TAS postal areas}
Only P-Q loads from 1\textsuperscript{st} January 2016 00:30 to 1\textsuperscript{st} June 2016 00:00 are considered to avoid any system instability near the beginning and the end of the outage period. Each load datum is 30 minutes apart, so in total \(24 hours \times 2 data/hour \times 152 days = 7296\) data are considered. 

Every Tasmanian region has a postal code. The data given in \cite{TasNetworks:load} include the substation name, which can be used to find the postal area in which the substation locates. A mapping of substation name to the located postal code would be constructed. Time series active and reactive load of individual postal areas are obtained using the constructed mapping. The load data from substations 'Huon River' and 'Newton' are not available for reasons of confidentiality. Their active and reactive loads would be taken as the average of the neighboring substations. Loads of 'Huon River' are taken as the average loads from substations 'Electrona', 'Kermandie', 'Kingston', 'Knights Road', 'Summerleas Zone'. Loads of 'Newton' are taken as the average loads from substations 'Queenstown', 'Rosebery', 'Savage River', 'Trial Harbour Zone'. 

\subsection{Assigning loads to P-Q buses}
\begin{figure}[htbp]
\centerline{\includegraphics[width=8cm]{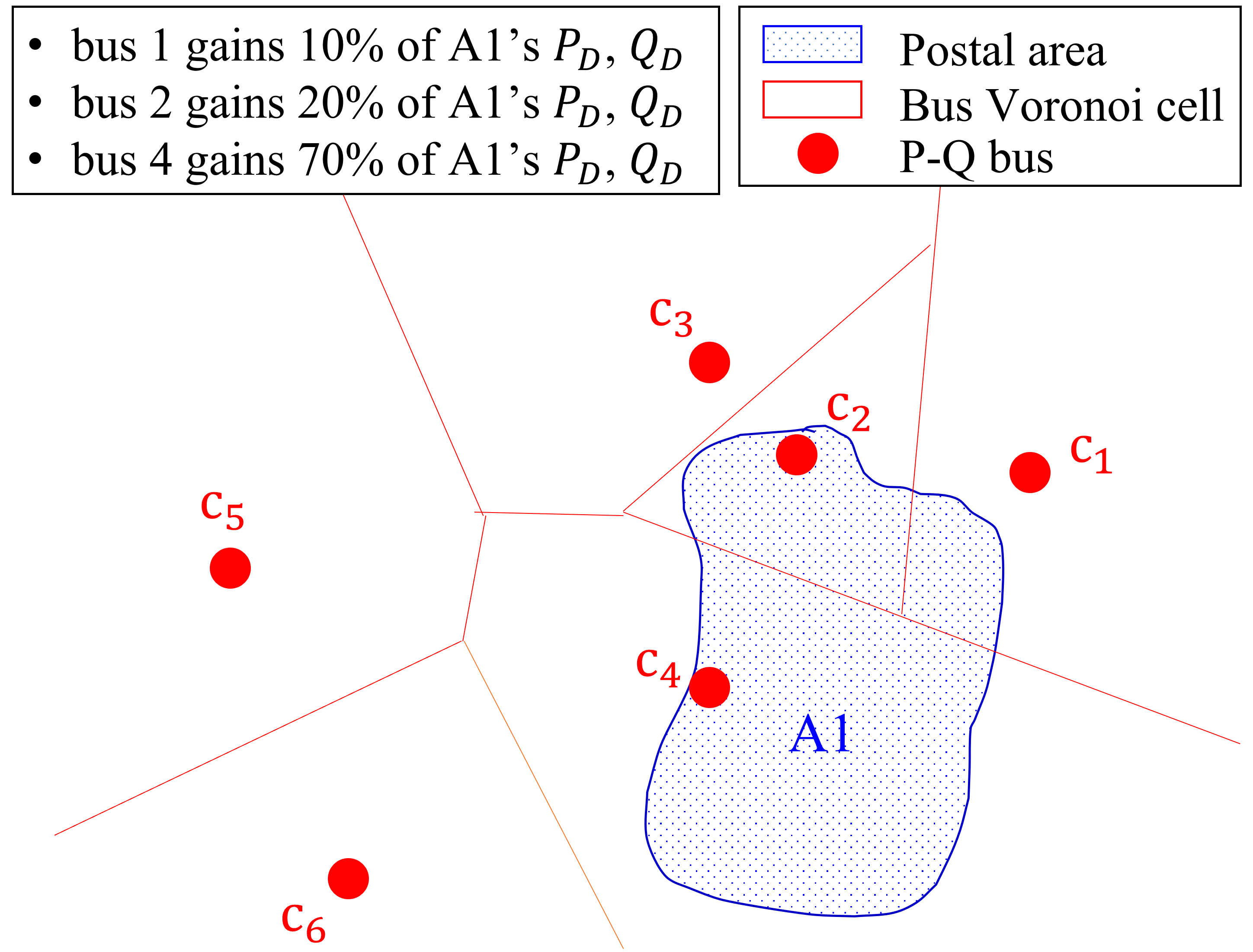}}
\caption{Example of assigning loads from one postal area to P-Q buses}
\label{figAlgorithm1illustration}
\end{figure}

The exact coordinates for each P-Q bus are known from \cite{Xenophon:2018}. Their locations are used to draw a Voronoi Diagram where each Voronoi cell corresponds to one P-Q bus. This Voronoi Diagram is overlayed on the Tasmanian postal area polygons \cite{ABS:2021}. For each P-Q bus cell \(c_i\) we calculate its intersection area with every postal area. Then, for each postal area that has an intersection with \(c_i\), it gives out a portion of its active and reactive loads to \(c_i\) with the portion as the ratio of the intersection area over the overall postal area. The operation to assign active and reactive loads to one bus at one time point is given in Algorithm~\ref{algoLoadAssignment}. Fig~\ref{figAlgorithm1illustration} illustrates an example of assigning active and reactive loads from one postal area to the neighboring P-Q buses. After assignment, all P-Q loads are linearly scaled such that their sum matches the historical Tasmanian regional load from \cite{AEMO:demand}.

\begin{figure}[!t]
 \removelatexerror
  \begin{algorithm}[H]
   \caption{P-Q bus load assignment\label{algoLoadAssignment}}
   \SetKwInOut{Input}{input}
   \SetKwInOut{Output}{output}
   \Input{Voronoi cell $c_i$ for P-Q bus $i$,\\Postal area polygons $[A1,...,AK]$,\\Postal area active loads $[P_{A1},...,P_{AK}]$,\\Postal area reactive loads $[Q_{A1},...,Q_{AK}]$}
   \Output{$P_{D_i}, Q_{D_i}$}
   $P_{Di}:=0$; $Q_{Di}:=0$\;
   \For{$k := 1$ to $K$}
   {
     Assign intersection area $a_k:=c_i\cap A_k$\;
     Assign area ratio $r_k:=a_k/A_k$\;
     Update $P_{Di}:=P_{Di}+r_k P_{Ak}$\;
     Update $Q_{Di}:=Q_{Di}+r_k Q_{Ak}$\;
   }

  \end{algorithm}
\end{figure}

The TAS network has one wind P-V bus. Since wind generators are highly dependent on weather, predicting their optimal generation values has minimal practical value. Instead, it is regarded as a negative P-Q bus with its active loads set as the negative of its historical active generations, which can be obtained from \cite{AEMO:dispatch}. Its reactive generation is approximated by multiplying its active generation by a sample from the sample normal distribution of \(Q_{Di}/P_{Di}\) on the same time interval, where \(i\in \mathcal{N}\). As a negative P-Q bus, its reactive load is the negative of this reactive generation.

\begin{figure}[h!]
\centerline{\includegraphics[width=7cm]{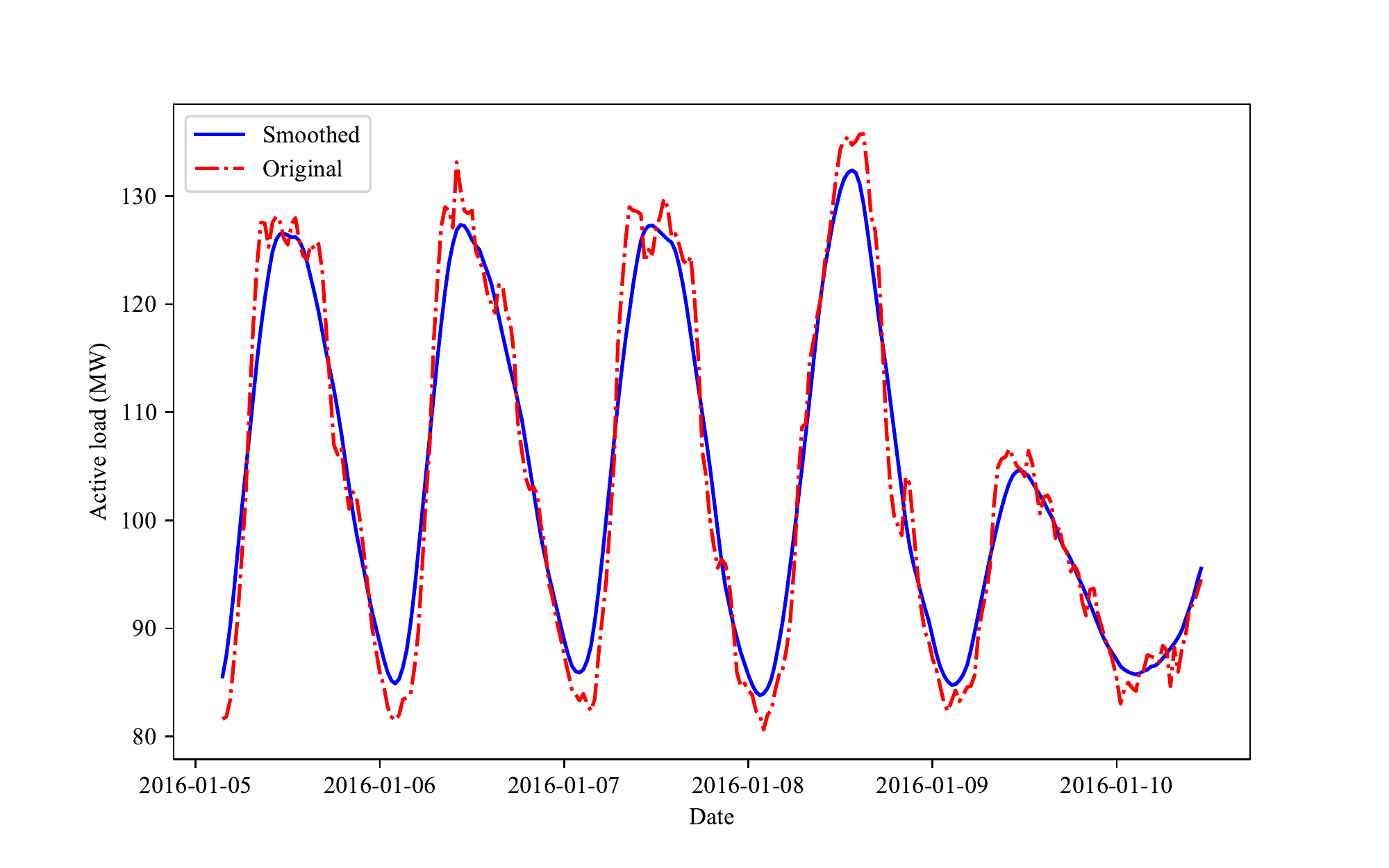}}
\caption{Active load of bus number 3}
\label{figActiveLoadBus3}
\end{figure}

Data smoothing is performed on the constructed time-series load data for each bus as a means to remove any possible outliers. A moving average of a window of 13 is applied to the data. \ref{figActiveLoadBus3} shows that the smoothed active load of bus number 3 maintains its overall shape in the time-series plot. 7284 realistic load data remains after the data smoothing.

\subsection{Calculating dispatch ground truths in TAS-97}
The ground truths of optimal dispatch at each network load data point are obtained using PYPOWER \cite{PYPOWER}. The TAS network topology information provided by \cite{Xenophon:2018} provides most information required for the AC-OPF formulation. Nonetheless, the dataset does not include the required information of generation cost functions, reactive generation limits, and branch flow limits. Their values are estimated as follows.

\subsubsection{Cost function for large hydroelectric generators}
Hydroelectric generators with active power capacity $P_{G_i}\geqq100MW$ would have linear cost functions of \(C_i= 6.724778 P_{G_i}\). This function references bus 30 in the pglib-opf test case \emph{pglib\_opf\_case39\_epri} \cite{pglib-opf}, which is a hydroelectric generator in a real power network in New England. 
\subsubsection{Cost function for small hydroelectric generators}
Hydroelectric generators with active power capacity $P_{G_i}<100MW$ would have linear cost functions of \(C_i= 10.087167 P_{G_i}\). This function is the cost function of larger hydroelectric generators scaled by 1.5, a factor taken from a report \cite{HEC:2018} written on hydroelectric power cost modeling in Tasmania.
\subsubsection{Cost function for natural gas generators}
For natural gas generators, their cost function are taken as the cost function from natural gas generators with similar capacity in the pglib-opf test case \emph{pglib\_opf\_case240\_pserc} \cite{pglib-opf}, which is the representation of a real network in the United States \cite{Munoz:2015}.
\subsubsection{Reactive generation limits}
For each generator $i$, the lower limit of reactive generation $Q_{Gi}^{\min}$ is taken as $Q_{Gi}^{\min}=-0.5P_{Gi}^{\max}$. The upper limit of reactive generation $Q_{Gi}^{\max}$ is taken as $Q_{Gi}^{\max}=0.5P_{Gi}^{\max}$. This estimation references the real-world pglib-opf test case \emph{pglib\_opf\_case240\_pserc} \cite{pglib-opf}. For bus 6, the upper and lower limits are doubled since it has been found that only in this arrangement PYPOWER can successfully solve all the load samples.
\subsubsection{Branch flow limits}
Branch flow limits are initially relaxed to high levels where branch flow constraint violations would be impossible. Then, the relaxed AC-OPF formulation and the realistic load samples are run with PYPOWER solver to obtain optimal dispatches under relaxed branch flow constraints. We assume that the branch flow limit $S_{ij}^{\max}$ for branch $ij$ increases with its maximum experienced flow in relaxed scenarios. Then, we set the branch flow limits using the following rule: if branch $ij$’s maximum experienced flow is the $k^{th}$ quantile among all maximum experienced flows, $S_{ij}^{\max}$ is set to the $k^{th}$ quantile of branch flow limits in the pglib-opf test case \emph{pglib\_opf\_case89\_pegase}. \emph{pglib\_opf\_case89\_pegase} is the representation of a small part of the European power network. This network is chosen for reference since it has a comparable size of buses and generation capacity to TAS-97.

P-V bus generation initial points for the PYPOWER OPF solver \emph{runopf} are set as the historical generation values from \cite{AEMO:dispatch}.

\subsection{Summary of TAS-97}
\begin{figure*}[htbp]
\centering
\includegraphics[width=\textwidth]{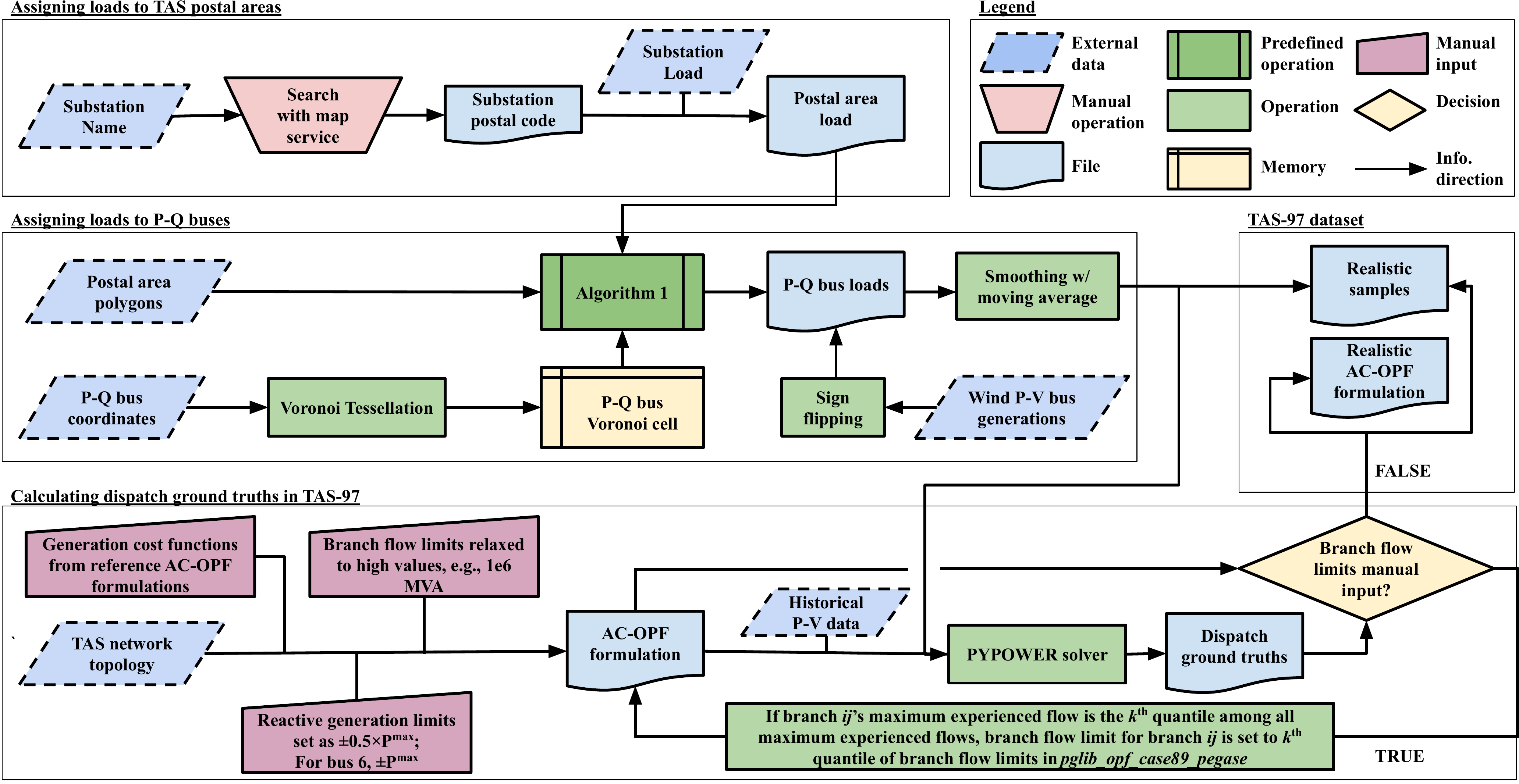}
\caption{Flowchart of the construction of TAS-97 dataset}
\label{figTAS97FlowChart}
\end{figure*}

\begin{figure}[htbp]
\centerline{\includegraphics[width=9.5cm]{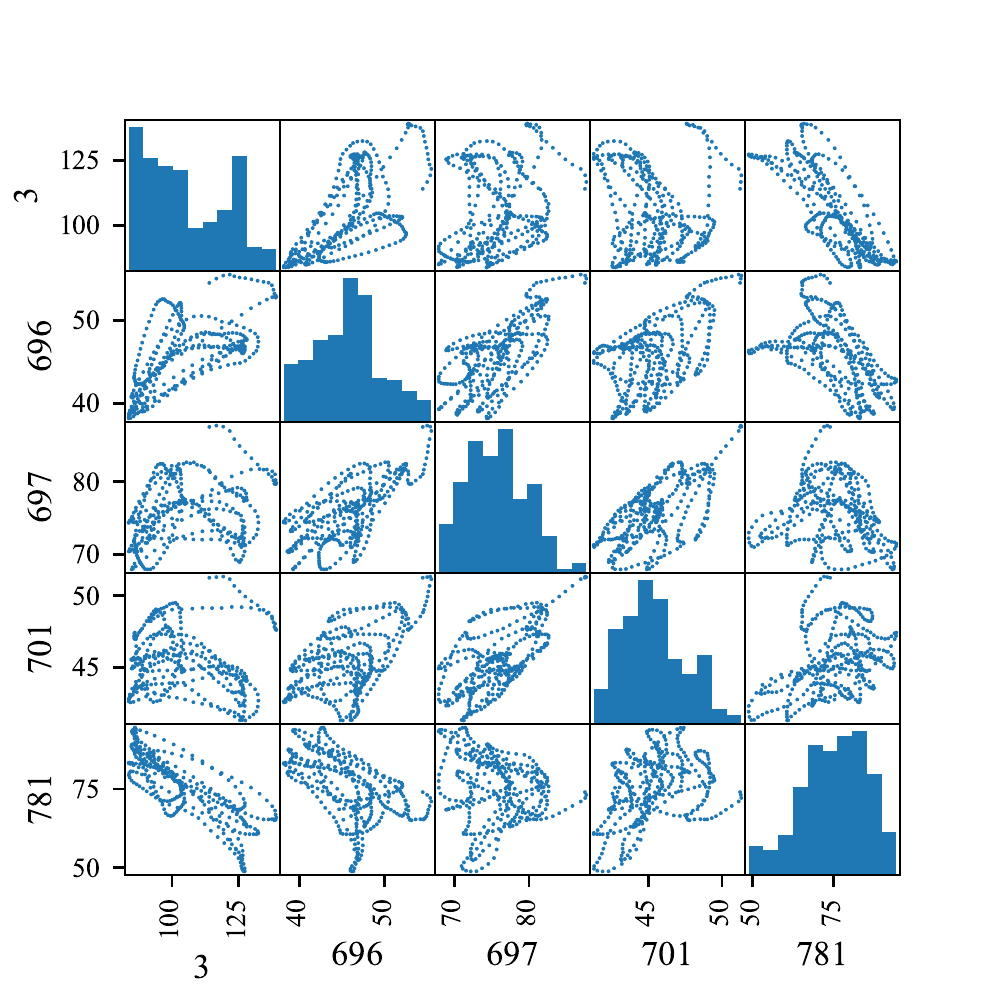}}
\caption{Scatterplot matrix of $P_{di}$ of selected P-Q buses in TAS-97}
\label{figCorrelatedPd}
\end{figure}

Fig.~\ref{figTAS97FlowChart} illustrates the construction process of TAS-97 dataset. It is an AC-OPF formulation-ready dataset that represents the Tasmanian power network from 1\textsuperscript{st} January 2016 03:30 to 31\textsuperscript{st} May 2016 21:00. The network has 78 P-Q buses, 19 P-V buses, and 126 branches. Generator types include hydroelectricity, natural gas, and wind. Accompanying this network topology is 7284 bus load-optimal dispatch data pairs. The P-Q bus loads are correlated. Fig.~\ref{figCorrelatedPd} shows positive correlation between bus 696 and 697 (correlation coefficient is 0.86), and negative correlation between bus 3 and 781 (correlation coefficient is -0.13). The dataset is available on GitHub \cite{TAS97GitHub}.

\section{Learning-based models}
\subsection{Model architecture}

Model architectures would be built upon the architecture DeepOPF of \cite{Pan:2021}, which showcases a feasibility-optimized end-to-end model that has high prediction accuracy, no violation rate, and two orders of magnitude of speedups on realistic networks and synthetic loads.

Define the following symbols:
\begin{tabular}{ll}
\emph{Symbol}&\emph{Definition}\\
\(P_{G0}\)&Active power generation on slack bus\\
\(Q_{G0}\)&Reactive power generation on slack bus\\
\(\lvert V_i\rvert\)&Voltage magnitude on bus $i$, $i \in \mathcal{D}$\\
\(\lvert V_0\rvert\)&Voltage magnitude on slack bus\\
\(\mathcal{G}'\)&Set of P-V buses that are not slack bus\\
\((\mathbf{P_D})_\mathcal{D}\)&Vector of $\{P_{Dj}:j\in\mathcal{D}\}$\\
\((\mathbf{Q_D})_\mathcal{D}\)&Vector of $\{Q_{Dj}:j\in\mathcal{D}\}$\\
\((\mathbf{P_G})_\mathcal{G'}\)&Vector of $\{P_{Gj}:j\in\mathcal{G}'\}$\\
\((\mathbf{Q_G})_\mathcal{G}\)&Vector of $\{Q_{Gj}:j\in\mathcal{G}\}$\\
\((\lvert \textbf{V}\rvert)_\mathcal{D}\)&Vector of $\{\lvert V_j\rvert:j\in\mathcal{D}\}$\\
\((\lvert \textbf{V}\rvert)_\mathcal{G}\)&Vector of $\{\lvert V_j\rvert:j\in\mathcal{G}\}$\\

\(\theta_i\)&Phase angle on bus \(i\)\\
\((\Theta)_\mathcal{D}\)&Vector of $\{\theta_{j}:j\in\mathcal{D}\}$\\
\((\Theta)_\mathcal{G'}\)&Vector of $\{\theta_{j}:j\in\mathcal{G'}\}$\\

$\mathbf{card}(X)$&Cardinality of set $X$\\
$\hat{y}$&Prediction of quantity $y$\\
$y^*$&Ground truth of quantity $y$\\

$\cdot \| \cdot$&Vector concatenation operator\\
\end{tabular}
\\

Our model is the zero-order optimized model proposed in \cite{Pan:2021}. The model consists of a feedforward neural network and a power-flow solver. The feedforward neural network $\mathcal{F}$ maps $(\mathbf{P_D})_\mathcal{D}\|(\mathbf{Q_D})_\mathcal{D}$ onto $(\lvert \textbf{V}\rvert)_\mathcal{G}\|(\mathbf{P_G})_\mathcal{G'}$. The feedforward process for each hidden layer is:
\begin{equation}
\label{eqnHiddenLayer}
\mathbf{h}_i(\mathbf{x}_i)=\sigma'(\mathbf{W}_i\mathbf{x}_i+\mathbf{b}_i)
\end{equation}
where $\sigma'(\cdot)$ is the ReLU activation function, $\mathbf{x}_i$ is the input vector to the hidden layer, $\mathbf{W}_i$ is the trainable connection weights, $\mathbf{b}_i$ is the trainable bias.
The network output ground truths are linearly scaled within their respective operating limits. The sigmoid activation function is adopted for the output layer such that the network outputs are always within the operating limits. The feedforward process of the output layer is:
\begin{equation}
\label{eqnOuputLayer}
\mathbf{o}(\mathbf{x}_o)=\sigma(\mathbf{W}_{o}\mathbf{x}_o+\mathbf{b}_{o})
\end{equation}
where $\sigma(\cdot)$ is the sigmoid activation function, $\mathbf{x}_o$ is the input vector to the last layer, $\mathbf{W}_{o}$ is the trainable connection weights, $\mathbf{b}_{o}$ is the trainable bias. Connected, the forward propagation of the neural network is:
\begin{equation}
\label{eqnF_A}
\mathcal{F}(\mathbf{P_D}, \mathbf{Q_D})=\mathbf{o}_(\mathbf{h}_{n}(\cdot\cdot\cdot\mathbf{h}_{2}(\mathbf{h}_{1}(\mathbf{P_D}\|\mathbf{Q_D}))))
\end{equation}
where $n$ is the number of hidden layers in the neural network.

\begin{figure}[htbp]
\centerline{\includegraphics[width=7.2cm]{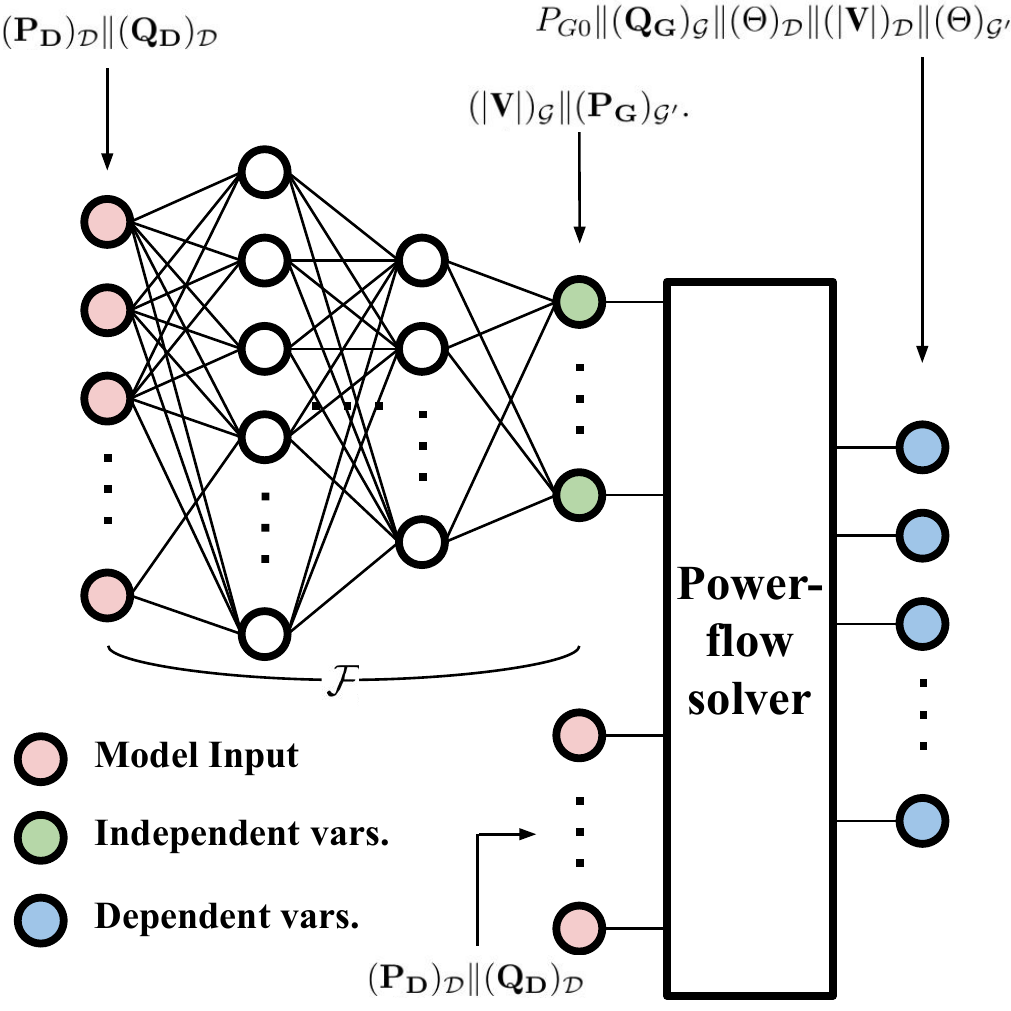}}
\caption{Model architecture}
\label{figModelA}
\end{figure}

The power-flow solver uses the neural network outputs as independent variables to calculate the dependent variables $P_{G0}\|(\mathbf{Q_G})_\mathcal{G}\|(\Theta)_\mathcal{D}\|(\lvert \textbf{V}\rvert)_\mathcal{D}\|(\Theta)_\mathcal{G'}$. PYPOWER's \emph{runpf} method would be used. Since the solver is based on Newton's method, different initial points for the dependent variables would affect the performance of the power-flow solver. The initial points are set as the average value of the dependent variables in the training data. Fig.~\ref{figModelA} illustrates the architecture of our model.

\subsection{Model training scheme}
In training the neural network in our model, the training scheme of \cite{Pan:2021} is adopted. That is, prediction loss $\mathcal{L}_{pred}$ from deviations between the neural network outputs and ground truths, as well as the penalty loss $\mathcal{L}_{pen}$ from operation constraints violations, are used. The prediction loss function is defined as the mean squared error:
\begin{equation}
\label{eqnLossPred}
\mathcal{L}_{pred}=(1/\mathbf{card}(\mathcal{F}))(\hat{\mathcal{F}}-\mathcal{F}^*)^T(\hat{\mathcal{F}}-\mathcal{F}^*)
\end{equation}
where $\hat{\mathcal{F}}$ is the prediction from $\mathcal{F}$, $\mathcal{F}^*$ is the vector of ground truths $(\mathbf{P_G})_\mathcal{G'}^*\|(\lvert \textbf{V}\rvert)_\mathcal{G}^*$ from a training sample. The gradients $\nabla\mathcal{L}_{pred}$ w.r.t. $\mathbf{W}_i$ and ${\mathbf{b}_i}$ are calculated by first computing the gradient w.r.t. $\hat{\mathcal{F}}$, then applying chain rule on each $\mathbf{W}_i$ and ${\mathbf{b}_i}$.

The penalty loss function is the deviation of the model prediction from the feasible region bounded by Equations (4) to (7). Let $y^{\max}$, $y^{\min}$ be the operation limits of quantity $y$, penalty loss on a prediciton $\hat{y}$ is
\begin{equation}
\label{eqnLossPenY}
p(\hat{y})=\max\{\hat{y}-y^{\max},0\}+\max\{y^{\max}-\hat{y},0\}
\end{equation}
and the total penalty loss on the prediction model is
\begin{equation}
\label{eqnLossPen}
\begin{aligned}
\mathcal{L}_{pen}=	&\frac{1}{\mathbf{card}(\mathcal{E})}\sum\nolimits_{(i,j)\in\mathcal{E}}p(\hat{S}_{ij})+	\\
				&\frac{1}{\mathbf{card}(\mathcal{D})}\sum\nolimits_{i\in\mathcal{D}}p(\lvert \hat{\textbf{V}}_i\rvert)+\\
				&\frac{1}{\mathbf{card}(\mathcal{G})}\sum\nolimits_{i\in\mathcal{G}}{p(\hat{Q}_{Gi})}+p(\hat{P}_{G0})+p(\hat{Q}_{G0}).
\end{aligned}
\end{equation}

$\mathcal{L}_{pen}$ does not depend on $\hat{\mathcal{F}}$, so a two-point zero-order optimization technique \cite{Agarwal:2010} is used to compute the gradient:
\begin{equation}
\label{eqnGradPen}
\hat{\nabla}\mathcal{L}_{pen}(\hat{\mathcal{F}})=\frac{\mathbf{card}(\mathcal{F})}{2\delta}\mathbf{v}(\mathcal{L}_{pen}(\hat{\mathcal{F}}+\mathbf{v}\delta)-\mathcal{L}_{pen}(\hat{\mathcal{F}}-\mathbf{v}\delta))
\end{equation}
where $\delta$ is a small smoothing parameter and $\mathbf{v}$ is a vector that has the same dimension as $\mathcal{F}$ and is randomly sampled on the unit ball.

\section{Experiments and results}
\subsection{Experimental setup}
Experiments are run on Windows 10 with quad-core (i7-10510U@1.80GHz) CPU with 8GB RAM. The model has $3$ hidden layers with $512,256,128$ hidden units. Both layer weight and bias initialization are sampled from the uniform distribution $\text{Unif}(-\sqrt{1/k},\sqrt{1/k})$, where $k$ is the number of hidden units in the layer. Training epochs is set to 200. The learning rate decays from $10^{-4}$ at epoch 1 to $10^{-9}$ at epoch 200. The learning rate is evenly spaced on the log scale from epoch 1 to epoch 200. Our empirical experience on TAS-97 shows including penalty gradient when the epoch is still low brings unstable penalty gradients that impair both the prediction accuracy and constraint satisfaction. Therefore the penalty gradient is only introduced after epoch 100. The penalty gradient is weighted by a 0.1 factor as suggested in \cite{Pan:2021}. 

To investigate the effect of sampling methods, we experiment on different sampling schemes.
\subsubsection{Testing on all 7284 realistic samples} We train three models on 65556 synthetic samples. Model 1 trains on samples synthesized from independent uniform distributions with bounds set as each bus’s historical load limits. Model 2 trains on samples synthesized from independent normal distributions fitted with each bus’s realistic loads. Model 3 trains on samples synthesized from a multivariate normal distribution fitted with all buses’ realistic loads.

\subsubsection{Testing on the last half of the realistic samples (3642 samples)} Experiments 4,5,6 use the models used in Experiments 1,2,3 respectively. Experiment 7 trains on the first half of the realistic samples (3642 samples).

Model performance is evaluated by the following metrics:
\begin{enumerate}
  \item Optimality loss $\eta_{opt}$: Average relative deviation between the model's and PYPOWER's dispatch cost.
  \item Feasibility rate $\eta_{fea}$: Percentage of feasible predictions of the model. A prediction is considered to be feasible if there are no constraint violation from Equations (4) to (7).
  \item Degree of deviation: Distances between a violated decision variable and the boundary in per-unit system for Equations (4) to (7). $\Delta_{P_G}$, $\Delta_{Q_G}$, $\Delta_{\lvert V\rvert}$, $\Delta_S$ denote the degree of violation of active generation, reactive generation, voltage magnitude, and branch flow respectively.
  \item Speedup factor $\eta_{sp}$: The average ratio of the prediction time $t_M$ taken by the model to the prediction time $t_0$ taken by PYPOWER.
\end{enumerate}

Table~\ref{tabTestingResults} shows the experimental results. 

\begin{table*}
\caption{Testing Results}
\begin{center}
\begin{tabular}{|c|c|c|c|c|c|c|c|c|c|c|c|c|c|}
\hline
\multirow{2}{*}{Experiment} & \multicolumn{2}{|c|}{Sampling scheme} &$\eta_{opt}$ & $\eta_{fea}$ & $\Delta_{P_G}$ & $\Delta_{Q_G}$ & $\Delta_{|V|}$ & $\Delta_{S}$ & $t_0$ & $t_M$ & \multirow{2}{*}{$\eta_{sp}$} \\
\cline{2-11}
& Training & Testing & (\%) & (\%) & (p.u.) & (p.u.) & (p.u.) & (p.u.) & (ms) & (ms) & \\
\hline
1 & 65556, synthetic (uniform) & 7284 & 2.89 & 94.10 & 0.014 & 0.000 & 0.000 & 0.000 & 855.51 & 20.26 & 42.23 \\
\hline
2 & 65556, synthetic (normal) & 7284 & 0.16 & 94.99 & 0.018 & 0.001 & 0.000 & 0.000 & 855.51 & 22.36 & 38.26 \\
\hline
3 & 65556, synthetic (multivariate normal) & 7284 & 0.13 & 99.73 & 0.0005 & 0.000 & 0.000 & 0.000 & 855.51 & 22.15 & 38.62 \\
\hline
4 & 65556, synthetic (uniform) & 3642 & 2.02 & 90.06 & 0.023 & 0.000 & 0.000 & 0.000 & 836.13 & 20.32 & 41.15 \\
\hline
5 & 65556, synthetic (normal) & 3642 & 0.99 & 92.30 & 0.027 & 0.001 & 0.000 & 0.000 & 836.13 & 17.55 & 47.64 \\
\hline
6 & 65556, synthetic (multivariate normal) & 3642 & 0.88 & 99.75 & 0.018 & 0.0004 & 0.000 & 0.000 & 836.13 & 22.69 & 36.85 \\
\hline
7 & 3642, realistic & 3642 & 2.58 & 94.73 & 0.029 & 0.003 & 0.000 & 0.000 & 836.13 & 26.80 & 31.20 \\
\hline
\end{tabular}
\end{center}
\label{tabTestingResults}
\end{table*}

\subsection{Effect of the quality of training samples}
Experiments 4 to 7 illustrate how the quality of training samples affects the model performance on realistic testing samples. The quality of training samples are listed below in descending order.
\subsubsection{Realistic training samples} While realistic training samples in Experiment 7 are expected to have the closest resemblance to the realistic testing samples, the model performance is not the best because of the limited training sample size. However, even with a small training size, the testing performance is better than Experiments 4 and 5, showcasing the quality of the samples. 

\subsubsection{Multivariate normal synthetic samples} Fig.~\ref{figCorrelatedPd} implies a multivariate normal probability distribution for the bus loads. Samples synthesized from a fitted multivariate normal distribution would have a close resemblance to the realistic testing samples, leading to the best tesing performance in Experiment 6. 

\subsubsection{Normal synthetic samples} Training samples from Experiment 5 disregard the correlation between bus loads but have a properly selected distribution family, therefore the model performance is worse than Experiment 6 but better than Experiment 4.

\subsubsection{Uniform synthetic samples} Training samples from Experiment 4 disregard both the correlation and have a poorly selected distribution family, therefore the model performance is the worst among Experiments 4 to 7.

\subsection{Learning-based OPF solver on realistic network and realistic loads}
Experiment 3 shows a promising result of using learning-based models on a realistic network and realistic loads. Both the optimality gap and degree of deviation are small, and the feasibility rate is high. 

There are violations on active generation constraints in Experiment 3. Note that the neural network’s sigmoid output activation function already forces the active generations from all generators except the slack generator to be within operational limits, therefore the degree of violation $\Delta_{P_G}$ comes only from the slack generator. With this small deviation, in practice we can make use of charging and discharging batteries in the slack generator substation to make sure the slack generator does not exceed its generation limits. 

\section{Conclusions}
We construct a dataset that represents the power network in Tasmania and included 7284 samples of realistic load data. It has been found that the realistic bus loads exhibit signs of a multivariate normal distribution. We train and test a feasibility-optimized end-to-end learning model using the dataset. Training on samples synthesized from a fitted multivariate normal load distribution, our model achieves 0.13\% cost optimality gap, 99.73\% feasibility rate, and 38.62 times of speedup on realistic testing samples when compared to PYPOWER.

\end{document}